\pdfoutput=1
\documentclass[11pt]{article}

\usepackage[preprint]{acl}

\usepackage{times}
\usepackage{latexsym}
\usepackage{amsfonts}
\usepackage[T1]{fontenc}
\usepackage[utf8]{inputenc}

\usepackage{microtype}

\usepackage{inconsolata}

\usepackage{graphicx}
\usepackage{booktabs}
\usepackage{multirow}
\title{Enhancing Depression Detection via Question-wise Modality Fusion}

\author{
 \textbf{Aishik Mandal\textsuperscript{1}},
 \textbf{Dana Atzil-Slonim\textsuperscript{2}},
 \textbf{Thamar Solorio\textsuperscript{3}},
 \textbf{Iryna Gurevych\textsuperscript{1}}
\\
 \textsuperscript{1}Ubiquitous Knowledge Processing Lab (UKP Lab)\\
Department of Computer Science and Hessian Center for AI (hessian.AI)\\
Technische Universität Darmstadt\\
 \textsuperscript{2}Department of Psychology, Bar-Ilan University\\
 \textsuperscript{3}MBZUAI
\\
 \small{\href{https://www.informatik.tu-darmstadt.de/ukp/ukp_home/index.en.jsp}{www.ukp.tu-darmstadt.de}
 }
 \\
}

\begin{document}
\maketitle
\begin{abstract}
Depression is a highly prevalent and disabling condition that incurs substantial personal and societal costs. Current depression diagnosis involves determining the depression severity of a person through self-reported questionnaires or interviews conducted by clinicians. This often leads to delayed treatment and involves substantial human resources. Thus, several works try to automate the process using multimodal data. However, they usually overlook the following: i) The variable contribution of each modality for each question in the questionnaire and ii) Using ordinal classification for the task. This results in sub-optimal fusion and training methods.
In this work, we propose a novel Question-wise Modality Fusion (\textit{QuestMF}) framework trained with a novel Imbalanced Ordinal Log-Loss (\textit{ImbOLL}) function to tackle these issues. The performance of our framework is comparable to the current state-of-the-art models on the E-DAIC dataset and enhances interpretability by predicting scores for each question. This will help clinicians identify an individual’s symptoms, allowing them to customise their interventions accordingly. We also make the code\footnote{\includegraphics[width=0.3cm]{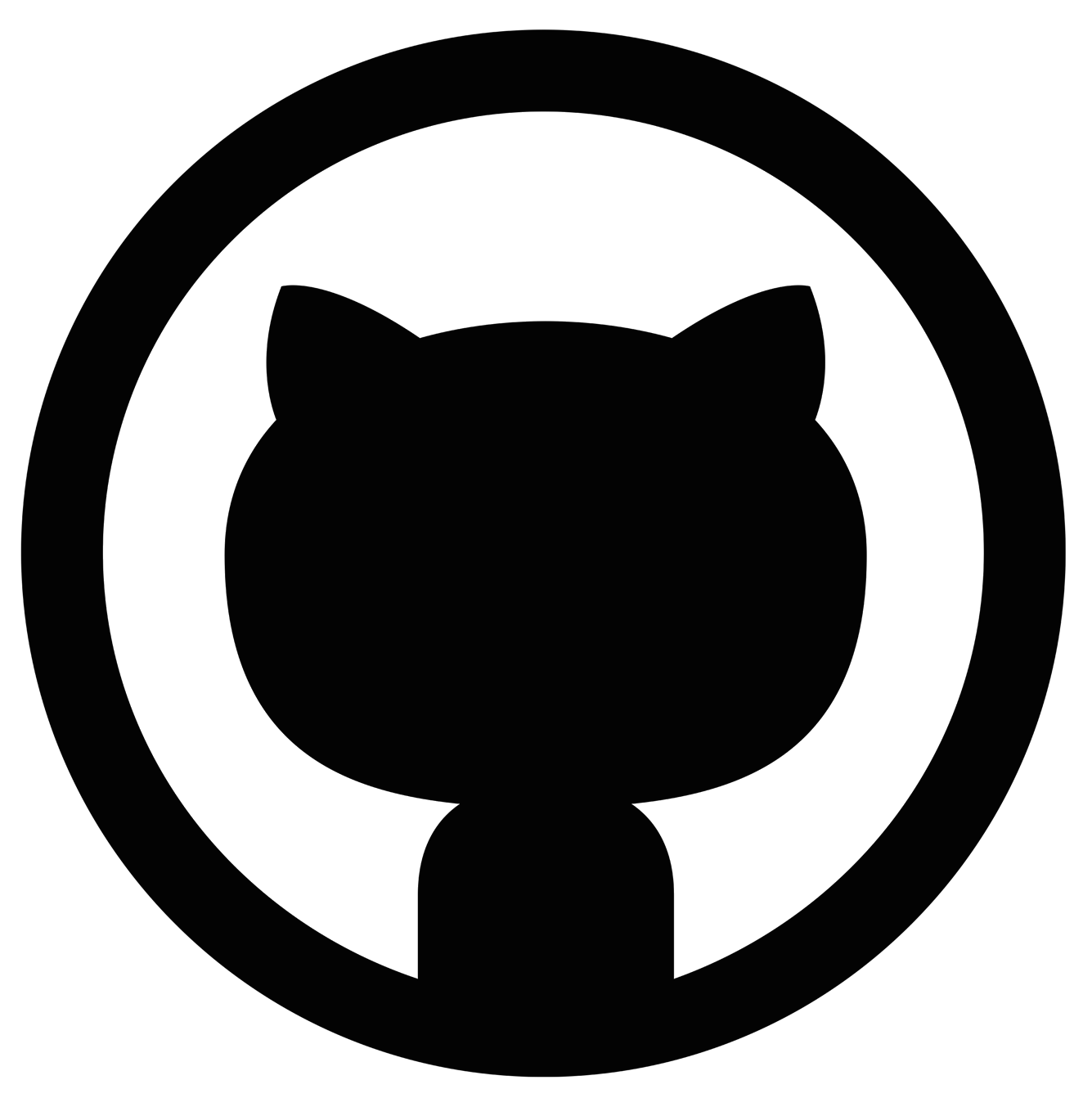} \href{https://github.com/UKPLab/clpsych2025-questmf}{QuestMF code}} for the \textit{QuestMF} framework publicly available.
\end{abstract}

\section{Introduction}

Depression is a major cause of disability globally \footnote{\href{https://www.who.int/news-room/fact-sheets/detail/depression}{WHO (2023, March 31). Depressive disorder.}}. Its personal and societal impact makes optimising mental health care practices crucial. Existing diagnostic systems of depression are heavily dependent on clinicians' proficiency in attending to patients' verbal and non-verbal cues, but achieving this expertise requires extensive training \cite{atzil2024leveraging}. The growing demand for mental health care services, coupled with a shortage of qualified providers, means that many individuals with depression go undiagnosed and untreated \cite{kazdin2021extending}.
Detection of depression severity is crucial, as it can prevent deterioration and enable adequate and effective treatment. Current diagnostic systems have faced criticism for failing to capture the significant heterogeneity and variability of symptoms between individuals \cite{bickman2020improving}. Understanding how different symptoms vary between individuals could lead to more personalised and effective interventions.

A common way to assess depression or track depression throughout a treatment program is based on self-reported questionnaires like PHQ-8 \citep{kroenke2009phq} or BDI-II \citep{beck1987beck}. These questionnaires contain questions regarding depression symptoms, and patients need to score each question based on how frequently they encounter these symptoms. The sum of the scores from each question gives the patient's depression severity score. However, such methods burden patients, especially when they are required to complete the questionnaires repeatedly as part of ongoing treatment monitoring \cite{kazdin2008evidence}. Thus, to improve the speed and convenience of diagnosing and monitoring depression, efforts are devoted to building depression severity prediction methods through machine learning. Initial works on automatic depression detection focused on using social media data \citep{LIWC-De-Choudhury} for binary depression classification. However, due to the lack of predicting depression severity, such a method is unable to prioritise people with higher levels of depression. So, the task was reformulated as classification among four depression levels \citep{naseem_ordinal}. However, models trained on social media data are unsuitable for clinical settings \citep{info:doi/10.2196/jmir.7215}.

Also, social media data lacks multimodal cues, which are often used by therapists to infer the depression severity of a patient. Depression has identifiable verbal and nonverbal characteristics, such as facial expression \citep{slonim2023facing}, prosodic information \citep{cummins2015review,5349358,SCHERER2014648, paz2024multimodal} and semantic features \citep{chim-etal-2024-overview}. To utilise these cues for depression assessment, the AVEC challenges \citep{avec_2017,avec_2019} released semi-clinical datasets, DAIC-WOZ \citep{gratch2014distress} and E-DAIC \citep{devault2014simsensei}, containing recorded interviews and self-reported PHQ-8 questionnaires. These questionnaires help detect symptoms and give more fine-grained depression severity levels.

Various works have tried to utilise the text, audio and video modality from the AVEC datasets and improve the fusion between them \citep{bert-gated-cnn,adaptive-fusion,cubemlp,sagan,multi-level-attention,yuan2022depression} to predict the depression severity score obtained from PHQ-8 questionnaires. However, these approaches only use one fusion module to fuse text, audio, and video information to predict the depression severity score (the sum of the scores for each question in the questionnaire). This design choice results in a failure to model the variable contributions of each modality depending on the questions in the questionnaire, leading to sub-optimal fusion.
For example, a question on being fidgety may require more attention to audio-visual modalities. On the other hand, text transcripts may contribute more significantly to a question regarding a person's appetite. While \citet{question-wise-text} also perform question-wise modeling, they mainly use text features concatenated with a few hand-crafted audio features from Automatic Speech Recognition (ASR), like sentiment, speech rate, and repetition rate. So, they do not utilise the audio and video modalities effectively.
Another issue is that the current multimodal methods frame the depression severity score prediction as a regression task, resulting in sub-optimal training. Humans score each question in the questionnaire as $0$, $1$, $2$, or $3$, depending on the frequency of the symptoms experienced. Thus, the depression severity prediction task should be framed as a question-wise ordinal classification task.

\textbf{Contributions:} We propose a novel Question-wise Modality Fusion (\textit{QuestMF}) framework for depression severity prediction. This framework contains question-wise fusion modules to ensure different contributions from modalities based on the question. In addition, we propose a novel Imbalanced Ordinal Log-Loss (\textit{ImbOLL}) function to train our models for ordinal classification. We find that our method matches the performance of the current state-of-the-art methods on the E-DAIC dataset and enhances interpretability for clinicians by identifying an individual's specific symptoms. We also analyse the importance of each modality for each question and find that a fusion of text and video modalities performs best in most questions.

\section{Background \& Related Work}
\label{sec:backgroung}

\subsection{Single Modality Methods}
\label{sec:related-works-unimodal}

Earlier works in depression severity prediction focused on the text modality like the use of linguistic feature extraction based on LIWC \citep{LIWC-De-Choudhury}, Bag-of-word models \citep{nadeem2016identifying}, word2vec embeddings \citep{husseini-orabi-etal-2018-deep} or using emotion features \citep{aragon-etal-2019-detecting}. With pre-trained language models like BERT \citep{devlin-etal-2019-bert} improving performances on text-based tasks, depression severity prediction works also utilised them \citep{bert-gated-cnn,temporal-cnn,cubemlp}. \citet{question-wise-text} introduces a framework to predict scores of each question of a PHQ-8 questionnaire to add interpretability, which is missing in the prior works.
These methods also ignore the multi-turn dialogue present in therapy sessions. Thus, \citet{milintsevich2023towards} introduces a turn-based method that encodes each dialogue turn using a sentence transformer \citep{reimers-gurevych-2019-sentence}. We use a similar turn-based model to encode the multi-turn dialogue data in each modality. We, however, use multihead self-attention instead of additive attention to improve the model.

With the advent of LLMs in recent times, \citet{sadeghi2023exploring} uses GPT-3.5-Turbo\footnote{\url{https://platform.openai.com/docs/models/gpt-3-5-turbo}} with encoder models for depression severity prediction. However, its performance falls short of the state-of-the-art models. Moreover, data privacy requirements do not allow data to be sent to proprietary LLMs. These issues motivate us to only explore encoder models.

The AVEC challenges introduced the potential to use audio features for depression detection. This resulted in works utilising low-level audio features \citep{Egemaps} extracted by OpenSmile \citep{opensmile}. \citet{temporal-cnn} uses CNNs over the low-level features, while \citet{mm-hierarchical} and \citet{cubemlp} use LSTMs to capture the temporal relation among them.
However, LSTMs are sub-optimal at processing long sequences. Thus, \citet{adaptive-fusion} uses transformers to process long sequences of audio features. There are also other methods that do not use OpenSmile features but rather use spectrograms \citep{bert-gated-cnn} or audio recordings directly \citep{han2023spatial,chen2023speechformer++}. However, they are computationally expensive, making them difficult to use for multimodal fusion. Here, we use LSTM over low-level features. We break the session into turns and aggregate features at the turn level to make shorter sequences that can be processed using LSTMs.

\subsection{Multimodal Fusion Methods}
\label{sec:related-works-mult}

Multimodal methods focus on improving the fusion of the modalities for depression severity prediction. \citet{question-wise-text} follows an early fusion method where they concatenate extracted text features with hand-crafted audio features. Other methods generally follow a late fusion strategy. Initial late fusion works used simple concatenation \citep{bert-gated-cnn} or weighted concatenation \citep{adaptive-fusion} for the fusion of text, audio, and video encodings. \citet{multi-level-attention} uses attention modules to improve fusion. Some works also use hierarchical fusion at frame level \citep{mm-hierarchical}, word level \citep{rohanian2019detecting}, or topic level \citep{guo2022topicattentive} to capture the interaction between the modalities at fine-grained levels. 
Fusion at the frame level, however, can cut words, and fusion at the word level lacks sufficient context, thus being too fine-grained. Meanwhile, fusion at the topic level is too course-grained. Thus, we use turn-level fusion in our framework. 
For further improvement in fusion, MMFF \citep{yuan2022depression} exploits the high-order interaction between different modalities. However, it is computationally expensive. CubeMLP \citep{cubemlp} uses MLPs to mix information among modalities to enhance the computational efficiency of fusion. However, it results in lower performance. \citet{sagan} uses Self-Attention GAN to augment training data to reduce the issue of data shortage. They use a cross-attention based fusion strategy \citep{Tsai2019MultimodalTF}. We use the same cross-attention-based fusion. However, these late fusion works use a single fusion module, thus ignoring the variable contribution of each modality according to the question. We use question-wise fusion modules to mitigate this issue.

\subsection{Ordinal Classification Methods}
\label{sec:ord-classification}
 
Ordinal classification has been explored in tasks like sentiment analysis in Twitter \citep{nakov2016semeval,rosenthal2017semeval}. While depression severity score prediction is also an ordinal classification task, prior multimodal methods \citep{bert-gated-cnn, milintsevich2023towards,multi-level-attention,yuan2022depression,sagan} treat it as a regression task. As a result, ordinal classification has been rarely explored in depression severity score prediction \citep{question-wise-text}. Ordinal classification methods include ordinal binary classification methods \citep{ord-bin-1,allwein2000reducing}, threshold methods \citep{ord-th-1,ord-th-2,ord-th-3} and loss-sensitive classification methods \citep{rennie2005loss,Diaz_2019_CVPR,Bertinetto_2020_CVPR,ord-th-3,castagnos-etal-2022-simple}. However, these methods are not suitable for imbalanced datasets. Since very few patients feel a specific symptom very frequently, the distribution of question-wise scores (labels) in depression severity prediction datasets is imbalanced. While \citet{question-wise-text} explores a weighted ordinal classification setup, their method can only be used with Kernel Extreme Learning Machine (KELM). Also, they obtain their best results with a regression framework and the results with weighted ordinal classification are far below state-of-the-art performance. Thus, we propose \textit{ImbOLL}, a modified version of the OLL function \citep{castagnos-etal-2022-simple} to handle the data imbalance, which can be used with any classification method.

In summary, our work is the first to perform question-wise modality fusion and present a loss function for imbalanced ordinal classification of depression severity prediction task that can be used with any classification method. Moreover, we are the first to analyse the contribution of each modality towards the score of each question, thus improving interpretability. We are also the first to show the robustness of our method through multiple seed runs.

\section{Dataset}
\label{sec:dataset}

Multimodal clinical data collection for depression detection is difficult due to privacy issues, thus resulting in small datasets. In this work, we use the E-DAIC dataset from the AVEC 2019 DDS \citep{avec_2019} challenge. While the E-DAIC dataset is also small, to the best of our knowledge, it is the only dataset with more than $200$ data points available for research on depression severity prediction. We do not use the DAIC-WoZ dataset since it is a subset of the E-DAIC dataset. Other available datasets are either smaller \citep{Zou2023SemiStructuralIC} or are not clinically grounded with self-reported questionnaires \citep{Yoon_Kang_Kim_Han_2022}.
The E-DAIC dataset collected recorded interview sessions with a virtual agent and filled out self-reported PHQ-8 questionnaires for each participant. All the interviews were conducted in English. The dataset provides text transcripts of participant dialogues, recorded audio clips, and visual features like ResNet, VGG, and OpenFace for each interview session. The recorded videos have not been released due to privacy concerns. The dataset contains $275$ sessions. The training set includes $163$ sessions, and the validation and test sets each contain $56$ sessions. However, one session from the validation and one from the test set have incomplete video feature files. Thus, we do not use them in the evaluation. The dataset also provides the PHQ-8 scores of all participants. The PHQ-8 score ranges from $0$ to $24$. While the training and validation sets contain scores for each of the eight PHQ-8 questions ($0$ to $3$), the test split only contains the total PHQ-8 scores. More details on the PHQ-8 questionnaire are provided in Appendix \ref{sec:appendix-phq8}.

\section{\textit{QuestMF} Framework}
\label{sec:methodology}

\begin{figure}[t]
  \includegraphics[width=0.9\columnwidth]{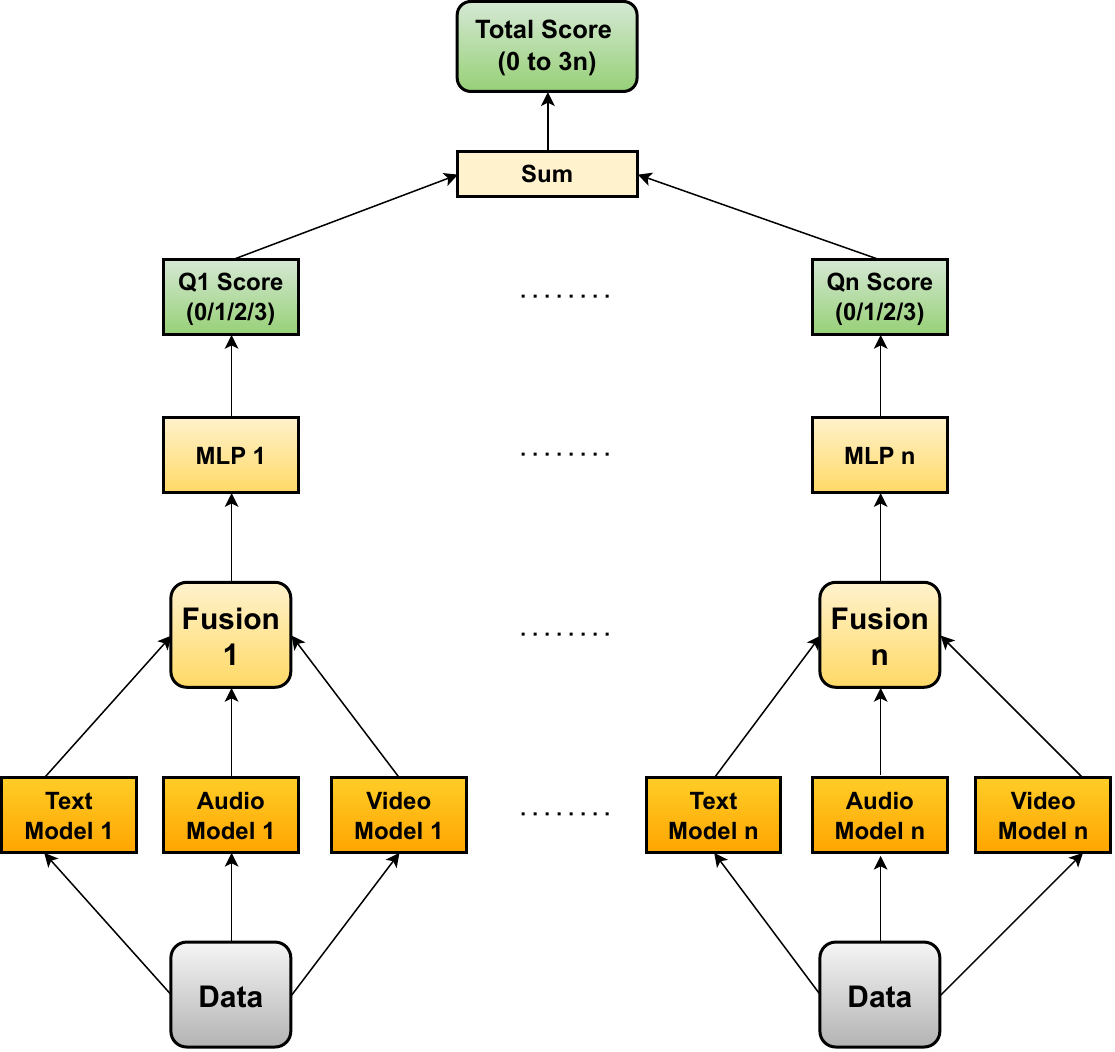}
  \caption{Proposed \textit{QuestMF} framework to predict depression severity score. Here, Qx denotes Question number x in the questionnaire. MLP denotes Multilayer Perceptron, which is used as the classification head. Each question is scored among classes $\{0,1,2,3\}$. These scores are then added to get the total score $\in \{0,1,2,...,3n\}$.}
  \label{fig:quest-wise-modeling}
\end{figure}

In this section, we present our novel Question-wise Modality Fusion (\textit{QuestMF}) framework. In this framework, we use $n$ different single modality encoders for each modality and $n$ different modality fusion models corresponding to $n$ questions (thus question-wise modality fusion) in a questionnaire. Each of the $n$ fused models outputs the score for its corresponding question ($0$, $1$, $2$, or $3$), ensuring different contributions from each modality depending on the question, which was lacking in previous works. These question-wise scores are added to get the total questionnaire score ($0$ to $3n$). Figure \ref{fig:quest-wise-modeling} shows the proposed framework. Moreover, the \textit{QuestMF} framework improves interpretability by predicting the score of each question. This allows clinicians to understand the symptoms affecting a patient and create interventions accordingly. This framework will also enable clinicians to track the progression of each symptom through the question-wise scores throughout the multiple therapy sessions during the treatment.

Moreover, current multimodal methods \citep{bert-gated-cnn, milintsevich2023towards,multi-level-attention,yuan2022depression,sagan} train the multimodal methods to predict the total depression severity score as a regression task treating it as a continuous variable. However, the question-wise scores belong to $4$ classes: $0,1,2,3$ depending on the frequency of the symptoms experienced, thus making it an ordinal classification task. While \citet{question-wise-text} tries this, they predominantly use text features and concatenate them with a few hand-crafted audio features. Also, they obtain their best result with regression setup, and their ordinal setup results are significantly worse than the state-of-the-art models. Treating question-wise scores as continuous variables also reduces interpretability as fractional scores like $1.5$ can mean experiencing a symptom at the frequency of either class $1$ or class $2$. Thus, framing the question-wise score prediction as an ordinal classification task ensures improved interpretability as we get the predicted probabilities of the $4$ classes: $0,1,2,3$ and choose the class with the highest probability. This is also more similar to how humans fill out the questionnaires.

Next, we discuss the single modality encoders used in the framework in Section \ref{sec:single-mod-enc} and the fusion methods used to combine the single modality encodings in Section \ref{sec:fusion-models}. Finally, we introduce the novel \textit{ImbOLL} function used to train the models for ordinal classification in Section \ref{sec:imboll-loss}.

\subsection{Single Modality Encoder Models}
\label{sec:single-mod-enc}

\begin{figure}[t]
  \centering
  \includegraphics[width=0.65\columnwidth]{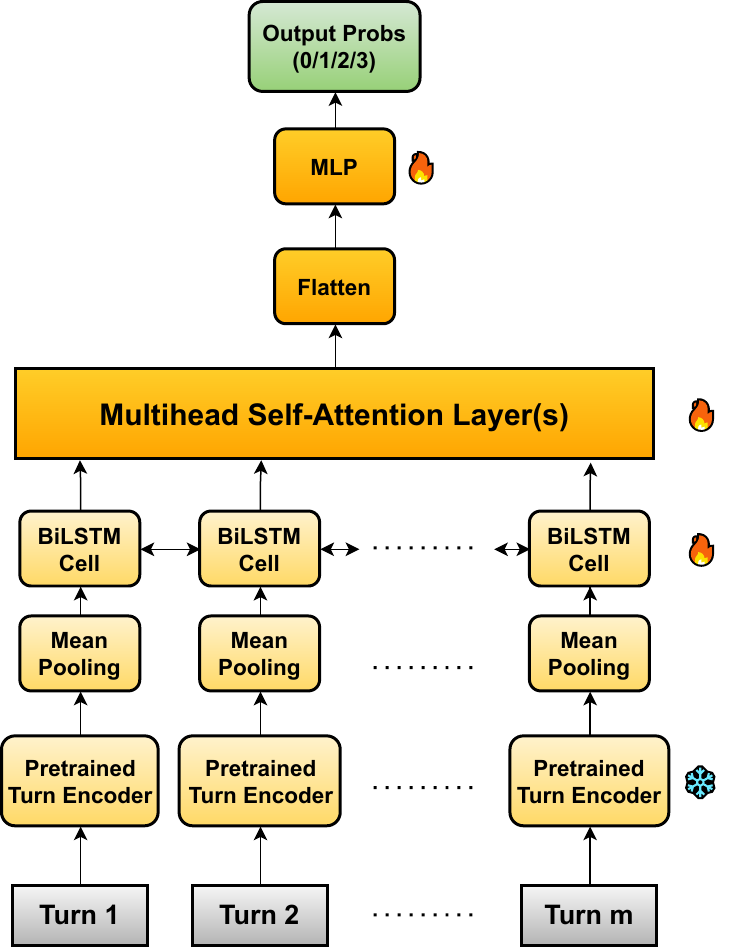}
  \caption{Architecture of single modality encoder models. We use a turn-based architecture to encode multi-turn dialogue data.}
  \label{fig:single-enc}
\end{figure}

All the single modality encoder models follow a turn-based method similar to \citet{milintsevich2023towards} to better encode the interviews containing multi-turn dialogues. We start by encoding the dialogue turns and then use these turn encodings to generate an encoding for the whole session. The overall structure is uniform across all modalities, as shown in Figure \ref{fig:single-enc}. Now, we describe each single modality encoder in detail.

\subsubsection{Text Encoder Model}

For text, we use the textual transcripts from interview sessions. We break the transcripts into dialogue turns and only consider the dialogue turns from participants. We encode the turns using a pre-trained sentence transformer \citep{reimers-gurevych-2019-sentence}. Now, we get turn encoding, $X_{i} \in \mathbb{R}^{tokens_{i} \times D_{model}}$, where $i \in (1,2,..,m)$ for $m$ turns, $tokens_{i}$ is the number of tokens in turn $i$ and $D_{model}$ is the model output dimension. Next, we use mean pooling over the tokens and normalise them to get $X_{mean, i} \in \mathbb{R}^{D_{model}}$. Then, we pass these turn encodings through a Bidirectional LSTM layer. The Bidirectional LSTM layer ensures that the turns can interact among themselves. This gives us $X_{lstm, i} \in \mathbb{R}^{2 \cdot D_{lstm}}$. Next, we use a multihead attention layer to determine the importance of each turn and get an updated encoding $X_{att, i} \in \mathbb{R}^{2 \cdot D_{lstm}}$. Following this, we concatenate and flatten the turn encodings obtained after the multihead attention layer to get a session-level encoding representation of $X_{session} \in \mathbb{R}^{(2m \cdot D_{lstm})}$. Finally, we pass this session-level encoding through a multilayer perceptron (MLP) to get the score probabilities.

\subsubsection{Audio Encoder Model}

For audio, we use the low-level features extracted by OpenSmile. Details on the low-level features extracted using OpenSmile are provided in Appendix \ref{sec:appendix-opensmile-features}. These features are extracted at every $0.01$ seconds. Like the text model, we process the information at the turn level. For each dialogue turn $i$, the dataset contains a starting time $t_{start, i}$ and an ending time $t_{end, i}$. We get the features extracted from time $t_{start, i}$ to $t_{end, i}$ and apply mean pooling to get the aggregated features in a turn, $X_{mean, i} \in \mathbb{R}^{D_{Features}}$.  Here, $D_{Features}$ is the number of the features extracted by OpenSmile. After this, we pass them through a Bi-LSTM layer and two attention layers to get updated turn encodings. These are concatenated, flattened, and passed through an MLP to get the score probabilities.

\subsubsection{Video Encoder Model}
\label{sub-sec:vid-enc}
For the video encoder model, we use ResNet \citep{He_2016_CVPR} features. Similar to text and audio, we aggregate information at turn level. For this, we get the ResNet features for the frames in a dialogue turn, i.e., frames in $t_{start, i}*sr:t_{end, i}*sr$, where $t_{start, i}$ is the starting time and $t_{end, i}$ is the ending time of the dialogue turn and $sr$ is the frames rate at which the video is recorded. We pass these ResNet features in a dialogue turn through a mean pooling layer and normalise them to get turn-level encoding $X_{mean, i} \in \mathbb{R}^{D_{ResNet}}$ where $D_{ResNet}$ is the output dimension of the ResNet model. After this, we follow the same architecture as the audio encoder to get the score probabilities.

\subsection{Modality Fused Models}
\label{sec:fusion-models}
For modality fusion, we use cross-attention based methods introduced by \citet{Tsai2019MultimodalTF}. The cross-attention layers are sometimes accompanied by X $\rightarrow$ Y. This denotes that the encoding of the Y modality is used as the query, and the encoding of the X modality is used as the key and value in the cross-attention layer. Next, we describe the modality fused models in detail.

\subsubsection{Two-Modality Fused Models}

\begin{figure}[t]
  \centering
  \includegraphics[width=0.65\columnwidth]{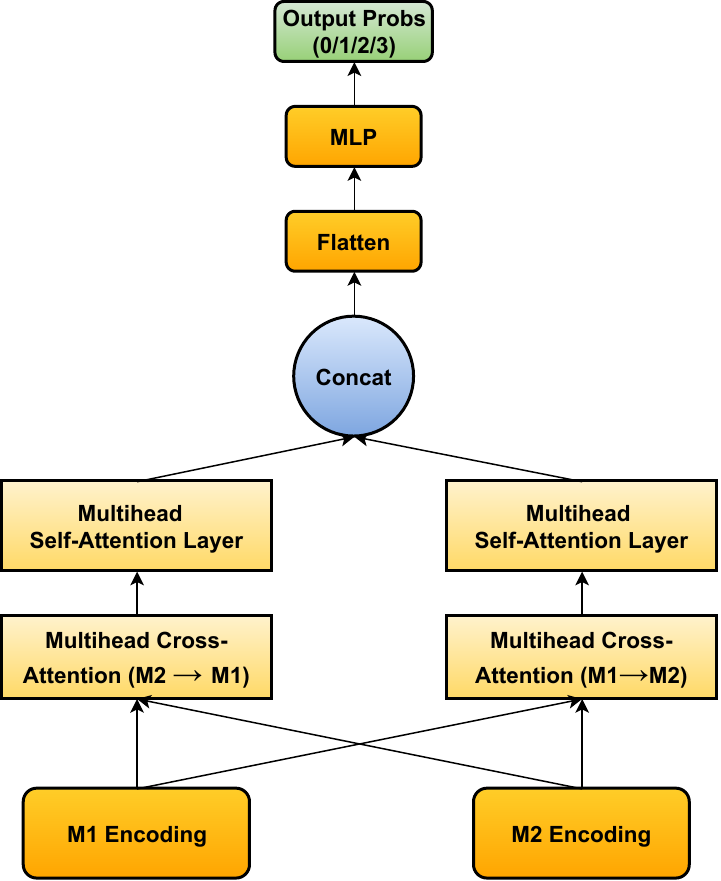}
  \caption{Architecture of two-modality fused models. We use cross-attention layers for interaction among modalities $M1$ and $M2$. In cross-attention, X $\rightarrow$ Y denotes that the Y modality encoding is used as the query and the X modality encoding as the key and value.}
  \label{fig:two-mod-fusion}
\end{figure}

Figure \ref{fig:two-mod-fusion} shows our two-modality fused models. We use the output from the multihead attention layers of the trained single modality encoders as the modality encoding. We use multihead cross-attention layers over these encodings to exchange information among the modalities. Considering modality encodings $M1$ and $M2$, we use two cross-attention layers $M1 \rightarrow M2$ and $M2 \rightarrow M1$ for interaction among the modalities. This is followed by a multihead self-attention layer for each cross-attention layer. Finally, we concatenate the encodings from the self-attention layers to get a fused encoding. We flatten this fused encoding and pass it through an MLP to get the score probabilities.

\subsubsection{Three-Modality Fused Models}

\begin{figure}[t]
  \includegraphics[width=\columnwidth]{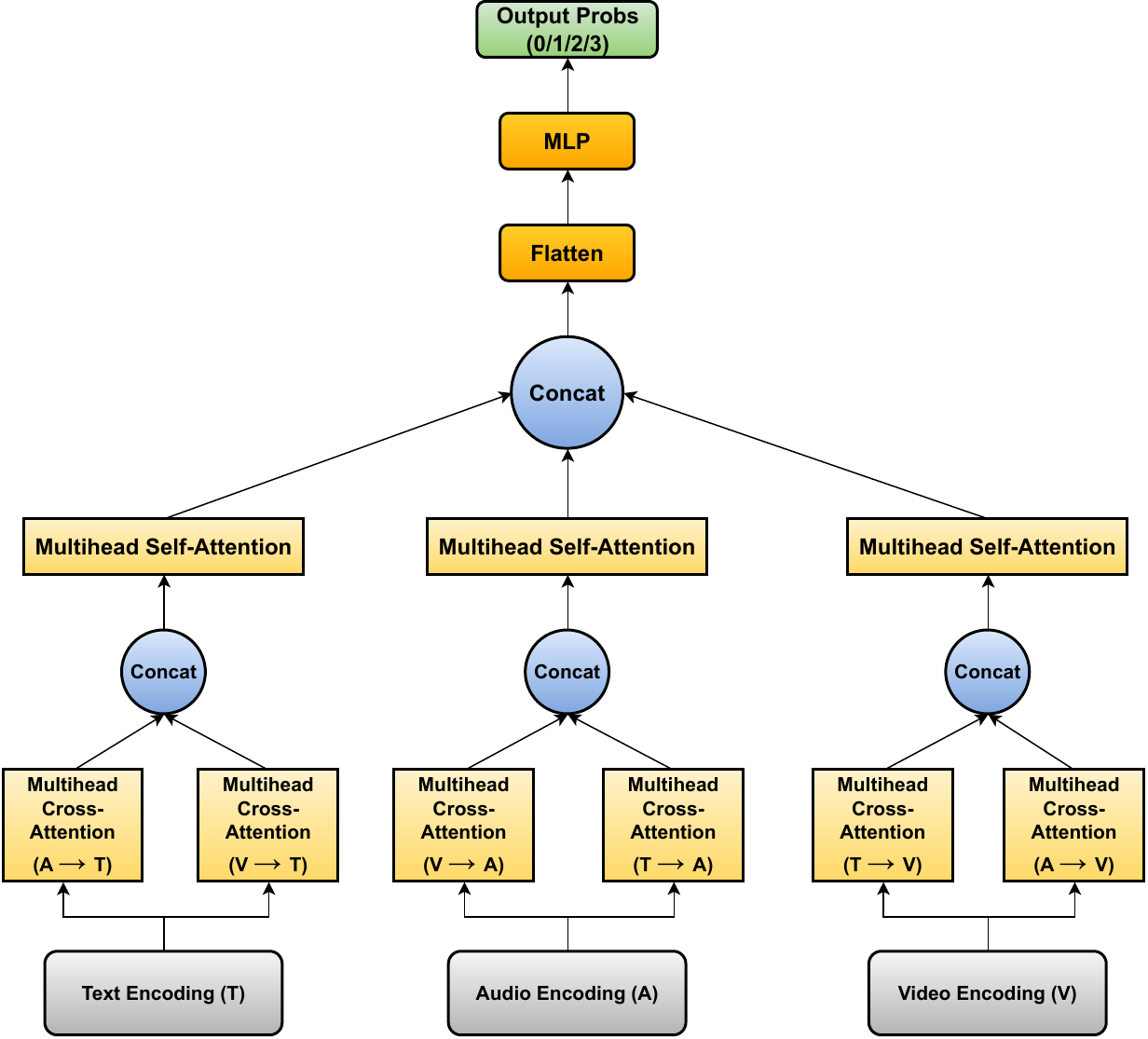}
  \caption{Architecture of the three-modality fused model. In cross-attention, X $\rightarrow$ Y denotes that the Y modality encoding is used as the query and the X modality encoding as the key and value.}
  \label{fig:three-mod-fusion}
\end{figure}

Figure \ref{fig:three-mod-fusion} shows our three-modality fused model. We use the output from the multihead attention layers of the trained single modality encoders as the modality encoding. Then, we use multihead cross-attention layers to pass information among the modalities. In this case, we have six combinations of query and (key, value) pairs.  
Now, we accumulate the encoding for each modality with information from the other two modalities. We perform this by concatenating the outputs from two cross-attention layers using the same modality as the query. For example, the audio encoding with information from text and video modalities will concatenate encodings obtained from cross-attention layers T $\rightarrow$ A and V $\rightarrow$ A. 
We pass these encodings for each modality with the information from other modalities through a multihead self-attention layer. Next, we concatenate them to get a combined encoding of all three modalities. Finally, we flatten the combined encoding and pass it through an MLP to get the score probabilities. 

\begin{table*}[t]
\centering
\begin{tabular}{lllll}
\toprule
Model                                                 & Modalities                & CCC($\uparrow$)   & RMSE($\downarrow$) & MAE($\downarrow$)  \\ \midrule
\citet{multi-level-attention}                                      & Text, Audio, Video        & 0.67  & 4.73 & 4.02 \\
\citet{cubemlp}                                             & Text, Audio, Video        & 0.583 & -    & 4.37 \\
\citet{sagan}                        & Text, Audio, Video & -     & \textbf{4.14} & \textbf{3.56} \\
\citet{yuan2022depression}                                                  & Text, Audio, Video        & \textbf{0.676} & 4.91 & 3.98 \\
\citet{question-wise-text} & Text, Audio                      & 0.62  & 6.06 & -    \\
\midrule
\textit{Total}                                    & Text, Audio, Video        & 0.618 & \textbf{4.99} & 4.03 \\
\textit{QuestMF} (MSE)                                        & Text, Audio, Video        & 0.620 & 5.31 & 4.16 \\
\textit{QuestMF} (OLL)                                        & Text, Audio, Video        & 0.656 & 5.17 & \textbf{3.89} \\
\textit{QuestMF} (\textit{ImbOLL})                                        & Text, Audio, Video        & \textbf{0.685} & 5.32 & 4.11 \\
\bottomrule
\end{tabular}
\caption{Results of \textit{QuestMF} trained with \textit{ImbOLL} function compared with ablation frameworks and prior works.}
\label{tab:main-results-table}
\end{table*}

\subsection{\textit{ImbOLL} Function}
\label{sec:imboll-loss}

Now, we introduce the novel \textit{ImbOLL} function we use to train our models. The \textit{ImbOLL} function is a modified version of the OLL \citep{castagnos-etal-2022-simple} function. The OLL function is used to train models for ordinal classification. The OLL function for $N$ classes is defined as:

\begin{equation} \label{eq:oll}
    \mathcal{L}_{OLL-\alpha}(P,y) = -\sum_{i=1}^{N}log(1-p_i)d(y,i)^{\alpha}
\end{equation}

where $y$ is the actual class, $p_i$ is the predicted probability of class $i$, $\alpha$ is a hyperparameter and $d(y,i)$ is the distance between the classes $y$ and $i$ which is defined as:

\begin{equation}
    d(y,i) = |y - i|
\end{equation}
The OLL function is based on the principle of penalising a model for bad decisions instead of rewarding good decisions. However, the OLL function is not suitable for imbalanced datasets. The questions of the PHQ-8 questionnaire consist of $4$ possible classes according to the frequency of symptoms: $0$, $1$, $2$, and $3$. However, very few participants give a score of $2$ or $3$ to a question as very few patients feel a particular symptom so often. This results in an imbalanced score distribution. To consider this, we introduce weights $w(y)$, which gives a harsher punishment to a model when it makes a wrong decision for a rarer ground truth score. The weights are defined as:

\begin{equation}
    w(y) = \frac{n_T}{n_y}
\end{equation}

Where $n_T$ is the total number of data points in the training set, and $n_y$ is the number of data points in the training set belonging to class $y$. Our novel loss function, \textit{ImbOLL}, is defined as follows:

\begin{equation} \label{eq:imboll}
    \mathcal{L}_{ImbOLL-\alpha,\beta} = -\sum_{i=1}^{N}log(1-p_i)d(y,i)^{\alpha}w(y)^{\beta} 
\end{equation}

where $\alpha$ and $\beta$ are hyperparameters.

\section{Experiments}
\label{sec:exps}

In this section, we describe the experiments with \textit{QuestMF} and its ablation frameworks. We use the following ablation frameworks:

\textbf{\textit{Total}:} We train the models with the MSE loss function to predict the total questionnaire score. This framework consists of a single modality encoder for each modality and a single fused model.

\textbf{\textit{QuestMF} (MSE):} We train the \textit{QuestMF} framework with the MSE loss function.

\textbf{\textit{QuestMF} (OLL):} We train the \textit{QuestMF} framework with OLL function.

\textbf{\textit{QuestMF} (\textit{ImbOLL}):} We train the \textit{QuestMF} framework with \textit{ImbOLL} function. \textit{This is our proposed framework}.

To evaluate the performance of the methods in the depression severity prediction task, we use the standard metrics used in prior works: Concordance Correlation Coefficient (CCC), Root Mean Squared Error (RMSE), and Mean Absolute Error (MAE). CCC is defined as:

\begin{equation}
      \label{eq:ccc}
      \rho_{c} = \frac{2 \rho \sigma_{x}\sigma_{y}}{\sigma_{x}^2 + \sigma_{y}^2 + (\mu_{x}-\mu_{y})^2}
\end{equation}
Where $\rho$ is the Pearson correlation between variables $x$ and $y$. $\sigma_{x}$ and $\sigma_{y}$ are the standard deviations of variables $x$ and $y$. $\mu_{x}$ and $\mu_{y}$ are the means of variables $x$ and $y$. We use CCC as the primary metric because it is unbiased by changes in scale and location \citep{ccc}. Psychologists use CCC to assess the agreement between test scores from different raters. It was also used as the evaluation metric for the AVEC 2019 challenge. We also report RMSE and MAE. A higher CCC score is desirable to show that the predicted and actual outputs correlate well. For RMSE and MAE, a lower value is desired, as it shows a smaller difference between the predicted and the actual output. More details on the metrics are provided in Appendix \ref{sec:appendix-metric}. For the \textit{ImbOLL} function, we empirically find the parameters $\alpha = 1$ and $\beta = 0.5$ to be the best. For the OLL function, we find $\alpha = 1$ to give the best results. The detailed results of the experiments with hyperparameters of \textit{ImbOLL} and OLL are presented in Appendix \ref{sec:appendix-imboll}. To show the robustness of our model, we run our experiments on three different seeds: $42$, $100$, and $1234$. We are the first in this domain to run experiments on multiple seeds. The training strategy, checkpoint selection, and hyperparameter details of the models are provided in Appendix \ref{sec:appendix-Exp-set-up}.

\begin{table*}[t]
\centering
\begin{tabular}{@{}llll@{}}
\toprule
Model   & CCC($\uparrow$) & RMSE($\downarrow$) & MAE($\downarrow$) \\ \midrule
\textit{Total}   &   $0.602 \pm 0.015$ &   $ \textbf{5.10} \pm \textbf{0.08}$   &  $3.99 \pm 0.05$   \\
\textit{QuestMF} (MSE) &   $0.602 \pm 0.024$  &  $5.36 \pm 0.24$  & $4.21 \pm 0.12$    \\
\textit{QuestMF} (OLL) &  $0.640 \pm 0.018$   &  $5.14 \pm 0.05$  &  $ \textbf{3.88} \pm \textbf{0.01}$   \\
\textit{QuestMF} (\textit{ImbOLL}) &  $ \textbf{0.662} \pm \textbf{0.022}$   &   $5.25 \pm 0.08$   &  $3.95 \pm 0.13$  \\ \bottomrule
\end{tabular}
\caption{Results of \textit{QuestMF} (\textit{ImbOLL}) framework over $3$ different seed runs compared with ablation frameworks.}
\label{tab:seed}
\end{table*}

\section{Results \& Analysis}
\label{sec:results}

The results comparing our proposed \textit{QuestMF} (\textit{ImbOLL}) framework with its ablations and current state-of-the-art methods are presented in Table \ref{tab:main-results-table}. Since the prior works only show their best results on a single run, we also pick our best results on CCC for a fair comparison. As we can see, our proposed \textit{QuestMF} (\textit{ImbOLL}) framework matches the performance of state-of-the-art models in CCC, the primary metric used to evaluate depression severity prediction tasks. In addition to comparable performance, \textit{QuestMF} framework provides question-wise scores, improving interpretability over current methods, thus allowing clinicians to design personalised interventions. We also show the robustness of our frameworks over multiple runs, which is not done by the previous works. The mean and standard deviation of the performance over $3$ runs are presented in Table \ref{tab:seed}. We observe that \textit{Total} and \textit{QuestMF} (MSE) perform similarly. However, training with ordinal classification objective improves the performance, as we can see from the results of \textit{QuestMF} (OLL). This shows the effectiveness of combining \textit{QuestMF} with ordinal classification training. Training with our novel \textit{ImbOLL} function further improves the results on CCC. It is also robust, with a standard deviation of $0.022$.

\begin{table}[!th]
\centering
\begin{tabular}{@{}llr@{}}
\toprule
\multicolumn{1}{c}{Modalities} & \multicolumn{1}{c}{Model} & \multicolumn{1}{c}{CCC($\uparrow$)}  \\ \midrule
\multirow{4}{*}{T}                 & \textit{Total}            & $0.591$ \\
                                      & \textit{QuestMF}(MSE)    & $0.593$ \\
                                      & \textit{QuestMF}(OLL)    & $0.592$ \\
                                      & \textit{QuestMF}(\textit{ImbOLL}) & $\textbf{0.615}$  \\ \midrule
\multirow{4}{*}{A}                & \textit{Total}            & $0.212$  \\
                                      & \textit{QuestMF}(MSE)    & $0.239$  \\
                                      & \textit{QuestMF}(OLL)    & $0.264$  \\
                                      & \textit{QuestMF}(\textit{ImbOLL}) & $\textbf{0.273}$  \\ \midrule
\multirow{4}{*}{V}                & \textit{Total}            & $-0.067$ \\
                                      & \textit{QuestMF}(MSE)    & $-0.075$  \\
                                      & \textit{QuestMF}(OLL)    & $\textbf{-0.041}$  \\
                                      & \textit{QuestMF} (\textit{ImbOLL}) & $-0.052$  \\ \midrule
\multirow{4}{*}{T+A}         & \textit{Total}            & $0.607$  \\
                                      & \textit{QuestMF}(MSE)    & $0.618$  \\
                                      & \textit{QuestMF}(OLL)    & $0.628$  \\
                                      & \textit{QuestMF}(\textit{ImbOLL}) & $\textbf{0.643}$ \\ \midrule
\multirow{4}{*}{T+V}         & \textit{Total}            & $0.610$  \\
                                      & \textit{QuestMF}(MSE)    & $0.627$ \\
                                      & \textit{QuestMF}(OLL)    & $0.628$  \\
                                      & \textit{QuestMF}(\textit{ImbOLL}) & $\textbf{0.659}$  \\ \midrule
\multirow{4}{*}{A+V}        & \textit{Total}            & $0.058$  \\
                                      & \textit{QuestMF}(MSE)    & $0.070$ \\
                                      & \textit{QuestMF}(OLL)    & $0.139$  \\
                                      & \textit{QuestMF}(\textit{ImbOLL}) & $\textbf{0.159}$ \\ \midrule
\multirow{4}{*}{T+A+V} & \textit{Total}            & $0.602$  \\
                                      & \textit{QuestMF}(MSE)    & $0.602$  \\
                                      & \textit{QuestMF}(OLL)    & $0.640$ \\
                                      & \textit{QuestMF}(\textit{ImbOLL}) & $\textbf{0.662}$ \\ \bottomrule
\end{tabular}
\caption{Ablation results for using different combinations of modalities with different frameworks. Here, T refers to Text, A refers to Audio, and V refers to Video. An addition between the modalities denotes using a fusion of them. The CCC scores presented are the mean over $3$ different seed runs.}
\label{tab:ablation-results-table}
\end{table}

\begin{figure*}[!th]
  \centering
  \includegraphics[width=0.9\linewidth]{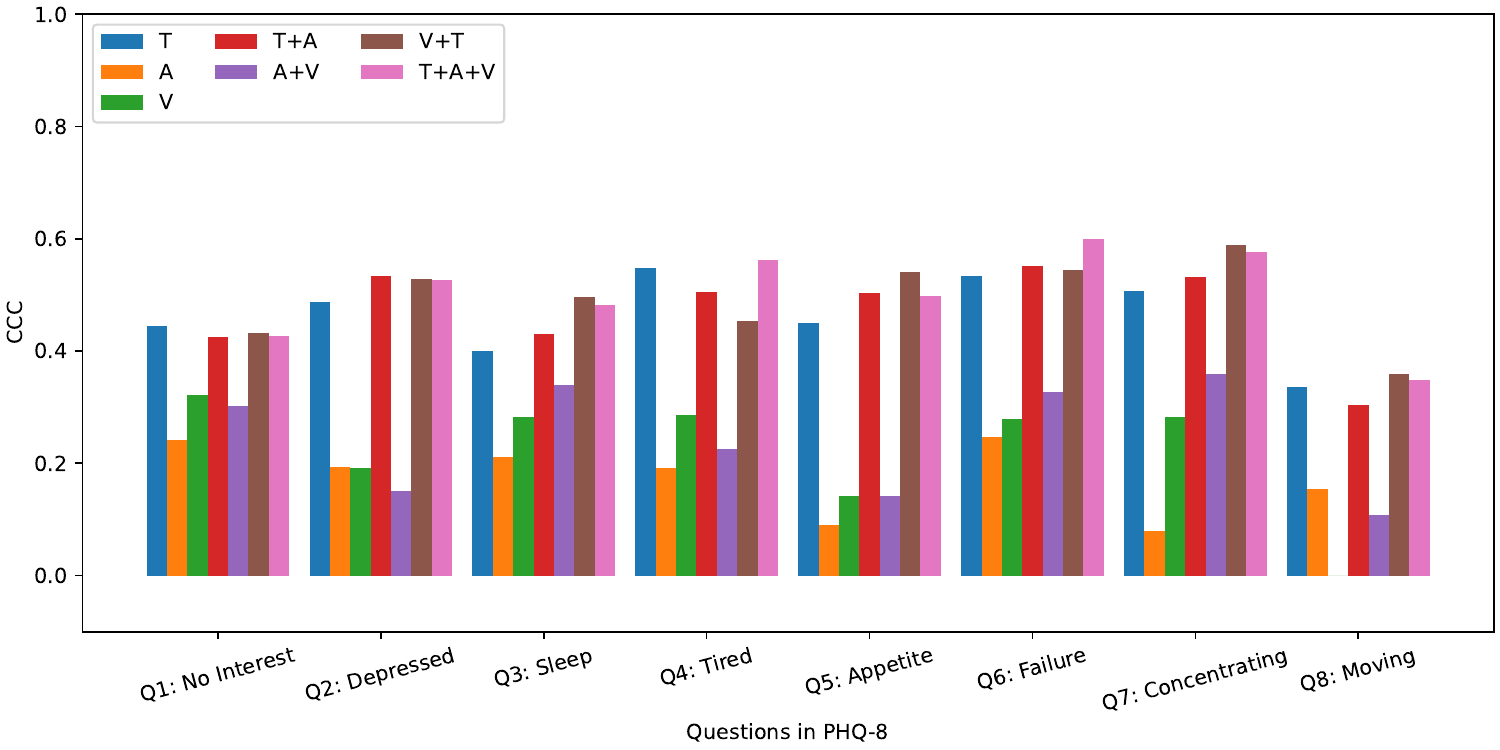} \hfill
  \caption {Validation CCC for each question with different modality models. Here, T refers to Text, A refers to Audio, and V refers to Video. An addition between the modalities denotes using a fusion of them. The video model for question $8$ gives the same scores to all data points. Thus, its CCC is not valid and is not shown in the graph.}
  \label{fig:ablation_quest_modality}
\end{figure*}

Next, we present an ablation study to observe the performance of single modality models, two-modality fused models, and the three-modality fused model with the different frameworks. We present the results in Table \ref{tab:ablation-results-table}. For all single modality models and two-modality fused models, we observe that \textit{QuestMF} (OLL) and \textit{QuestMF} (\textit{ImbOLL}) frameworks show better performance than the other frameworks. All the frameworks except \textit{QuestMF} (\textit{ImbOLL}) and \textit{QuestMF} (OLL) show the best performance with only text and video fusion, while the performance drops when all three modalities are fused. This shows that training with an ordinal classification task gives a better optimisation objective. Moreover, we also see that \textit{QuestMF} (\textit{ImbOLL}) and \textit{QuestMF} (OLL) show the best performance gains when adding more modalities. Comparing the results of Text + Audio + Video with the Text models, we observe that \textit{QuestMF} (\textit{ImbOLL}) achieves an improvement of $0.047$ (Text + Audio + Video ($0.662$) $-$ Text ($0.615$)) on CCC and \textit{QuestMF} (OLL) achieves an improvement of $0.048$ (Text + Audio + Video ($0.640$) $-$ Text ($0.592$)). Among the regression methods, \textit{Total} achieves the best improvement when comparing the results of Text + Audio + Video with the Text models. However, it only achieves an improvement of $0.011$ (Text + Audio + Video ($0.602$) $-$ Text($0.591$)). This further shows that the \textit{QuestMF} (\textit{ImbOLL}) framework improves fusion. Appendix \ref{sec:appendix-ablation} provides more detailed results for this ablation.

Finally, we analyse the importance of each modality toward predicting the score for each question in the framework.
This analysis can be used to build a framework where each question only uses the modalities important for its scoring. However, we do not build such a framework as it requires more data to generalise the analysis.
Since we lack fine-grained labels in the test split, we use the validation set CCC for each question in this analysis. A higher CCC shows greater importance. The CCC for each question with different modality combinations is shown in Figure \ref{fig:ablation_quest_modality}. From them, we observe:

\textbf{Q1: Feeling no interest.} Text modality gives the best results. This may be because the content of the conversation in the interview is the best indicator to determine loss of interest in hobbies.

\textbf{Q2: Feeling depressed.} Text + Audio gives the best results with very close results from Text + Video and Text + Audio + Video. Audio cues like flat speech and visual cues like sadness or blunted facial expressions might help predict this score. However, a fusion of all three struggles to train due to a small number of training data.

\textbf{Q3: Irregular sleep.} Text + Video gives the best results. Sleep disturbances are often directly reported by individuals, which explains the contribution of the text modality. The video modality contributes since sleep issues can often be observed from posture and general demeanour.

\textbf{Q4: Feeling tired.} Text + Audio + Video gives the best results as tiredness is often visible in a person's face, body language, and way of speaking.

\textbf{Q5: Irregular appetite.} Text + Video gives the best results. A person's appetite can be determined by directly asking them, so the text has the highest importance here. The video also contributes as physical appearance may influence the prediction.

\textbf{Q6: Feeling like a failure.} Text + Audio + Video gives the best results. Acoustic cues like a disappointed voice and visual cues like a saddened face help in the prediction.

\textbf{Q7: Trouble concentrating.} Text + video gives the best results. Visual cues like gaze can be an important factor. Looking away and not making eye contact may indicate concentration problems.

\textbf{Q8: Irregularities in moving and speaking.} In this case, Text + Video gives the best results, with Text + Audio + Video closely following. Movement can be captured through the video, and irregularities in a speech can be captured from transcripts and audio recordings.

\section{Conclusions \& Future Work}
\label{sec:conclusion}
In this work, we show that our Question-wise Modality Fusion (\textit{QuestMF}) framework trained with Imbalanced Ordinal Log-Loss (\textit{ImbOLL}) function improves the interpretability in depression severity score prediction by predicting scores of specific questions. This can help clinicians identify particular symptoms or symptom combinations, enabling them to tailor their interventions to the individual's specific needs. The \textit{QuestMF} (\textit{ImbOLL}) framework also shows performance comparable to current state-of-the-art models on the E-DAIC dataset. We also show its robustness over different seeds. Our framework can assist clinicians in diagnosing and monitoring depression and reduce the burden placed on patients in filling out self-reported questionnaires.
Additionally, we perform an extensive analysis to understand the importance of each modality for each question in the questionnaire. By releasing the code, we hope to enable future research of this framework on other questionnaires for mental health assessment and on real world longitudinal therapy data.

\section*{Limitations}

While the question-wise modality fusion framework trained with \textit{ImbOLL} function offers a solution to considering the variable contribution from modalities based on questions and framing the problem as an ordinal classification task, the data used for training and evaluation are not ideal. While the E-DAIC dataset was released to improve multimodal research in depression severity prediction, the training split only contains 163 sessions. As a result, the trained models are prone to overfitting and high bias and are unlikely to perform well in out-of-distribution data. The validation and test splits also contain only $56$ sessions each. As a result, they are far from representing the general population. Moreover, bigger and more diverse datasets are unavailable due to privacy issues. Thus, the \textit{QuestMF} (\textit{ImbOLL}) framework is only tested on the E-DAIC dataset, which further constrains testing the generalisability of the model. Thus, the results and analysis obtained in this work need to be verified with bigger and more diverse datasets in the future. Also, the E-DAIC dataset only contains first time interviews of participants with a virtual agent similar to enrolling interviews for therapy and does not contain real therapy session interviews. Real world depression tracking also requires longitudinal data, i.e., multiple therapy sessions with the same participant and tracking changes in depression severity throughout their treatment. Since the E-DAIC dataset does not contain such data, we cannot test the effectiveness of our model in such real world situations.

Another limitation is the language and culture coverage. In this work, we only cover the English language, and the dataset is collected in the US. However, people use different languages to express themselves, and people from different cultures express themselves differently, thus affecting therapy. However, \textit{QuestMF} (\textit{ImbOLL}) could not be developed and tested for such generalisation due to the lack of suitable datasets.

Our focus in this work is to present a more intuitive methodology that considers the variable contribution from modalities according to the question in a questionnaire, frames the task in its true nature of ordinal classification task, gives question-wise scores that can help clinicians design more personalised interventions and analyse the results to understand the contribution of each modality towards the score of each question.

\section*{Ethical Considerations}

While this work is focused on presenting a methodology and analysis for automatic depression detection, the methods need to be trained on larger datasets to ensure the method's generalisation capabilities. The method should also be assessed for generalisability through clinical trials. Deploying these methods without proper training and assessment through clinical trials could lead to introducing harmful biases in real world situation. Therefore, the framework trained with the E-DAIC dataset may not be used in clinical practice. It requires a broader evaluation and clinical validation before being used in real-world clinical settings.

\section*{Acknowledgements}
This research work has been funded by the German Federal Ministry of Education and Research and the Hessian Ministry of Higher Education, Research, Science and the Arts within their joint support of the National Research Center for Applied Cybersecurity ATHENE. This work has also been supported by the DYNAMIC center, which is funded by the LOEWE program of the Hessian Ministry of Science and Arts (Grant Number: LOEWE/1/16/519/03/09.001(0009)/98).
\bibliography{custom}

\appendix

\section{PHQ-8 Questionnaire}
\label{sec:appendix-phq8}
In this section, we provide more details regarding the PHQ-8 questionnaire. The PHQ-8 questionnaire consists of the following questions:

\begin{itemize}
    \item \textbf{Question 1:} Little interest or pleasure in doing things.
    \item \textbf{Question 2:} Feeling down, depressed, irritable or hopeless.
    \item \textbf{Question 3:} Trouble falling or staying asleep, or sleeping too much.
    \item \textbf{Question 4:} Feeling tired or having little energy.
    \item \textbf{Question 5:} Poor appetite or overeating.
    \item \textbf{Question 6:} Feeling bad about yourself – or that you are a failure or have let yourself or your family down.
    \item \textbf{Question 7:} Trouble concentrating on things, such as school work, reading or watching television.
    \item \textbf{Question 8:} Moving or speaking so slowly that other people could have noticed? Or the opposite – being so fidgety or restless that you have been moving around a lot more than usual.
\end{itemize}

These questions are scored from $0$ to $3$ based on how frequently the patients encounter them in the last two weeks. The scoring is based on the following:

\begin{itemize}
    \item \textbf{Score 0:} Not at all
    \item \textbf{Score 1:} Several Days
    \item \textbf{Score 2:} More than Half Days
    \item \textbf{Score 3:} Nearly Everyday
\end{itemize}

The total score from all the questions is used to determine the depression severity of a patient. A higher score denotes higher depression severity.

\section{Evaluation Metrics}
\label{sec:appendix-metric}

We use three different evaluation metrics for evaluation in this paper, which are elaborated below:

\begin{itemize}
    \item \textbf{Concordance Correlation Coefficient (CCC):} CCC is a correlation based metric. It varies from $-1$ to $1$. A CCC of $-1$ between predicted values and actual values means the two variables are opposite. A CCC of $1$ means they are identical, and $0$ means they are not correlated. Thus, a higher CCC is desirable. CCC is defined as follows:
    \begin{equation}
      \label{eq:ccc2}
      \rho_{c} = \frac{2 \rho \sigma_{x}\sigma_{y}}{\sigma_{x}^2 + \sigma_{y}^2 + (\mu_{x}-\mu_{y})^2}
    \end{equation}
    where $\rho$ is the pearson correlation between variables $x$ and $y$. $\sigma_{x}$ and $\sigma_{y}$ are the standard deviations of variables $x$ and $y$. $\mu_{x}$ and $\mu_{y}$ are the means of variables $x$ and $y$.

    \item \textbf{Root Mean Squared Error (RMSE):} RMSE is a standard metric used in regression problems. It varies from $0$ to $\infty$. A lower RMSE is desirable. It is defined as follows:
    \begin{equation}
      \label{eq:rmse}
      RMSE = \sqrt{\frac{1}{N}\sum_{i=1}^{N} \| y(i) - x(i)\|^2}
    \end{equation}
    \item \textbf{Mean Absolute Error (MAE):} Mean Absolute error is another standard metric used in the evaluation of regression problems. It varies from $0$ to $\infty$. It is defined as follows:
    \begin{equation}
      \label{eq:mae}
      MAE = \frac{1}{N}\sum_{i=1}^{N} \| y(i) - x(i)\|
    \end{equation}
\end{itemize}

\section{OpenSmile Low-level Features}
\label{sec:appendix-opensmile-features}
Here, we provide more details about the low-level features extracted by OpenSmile, which are used in our experiments.
\subsection{Frequency related parameters}

\textbf{Pitch}: Logarithmic F0 on a semitone frequency scale, starting at 27.5 Hz. A semitone is the smallest music interval and is considered the most dissonant when sounded harmonically.

\textbf{Jitter}: Jitter measures the cycle-to-cycle variations of the fundamental frequency. It is a measure of frequency variability compared to the person’s fundamental frequency.

\textbf{Formant 1, 2, and 3 frequency}: Centre frequency of first, second, and third formant. Formants are distinctive frequency components of the acoustic signal produced by speech. They are used to identify vowels.

\textbf{Formant 1 Bandwidth}: Bandwidth of the first formant. The formant bandwidth affects the identification of vowels in competition with other vowels.

\subsection{Energy/Amplitude related parameters}

\textbf{Shimmer}: Shimmer measures the cycle-to-cycle variations of fundamental amplitude. The shimmer changes with the reduction of glottal resistance and mass lesions on the vocal cords and is correlated with the presence of noise emission and breathiness.

\textbf{Loudness}: an estimate of perceived signal intensity from an auditory spectrum. 

\textbf{Harmonics-to-noise ratio}: Relation of energy in harmonic components to energy in noise-like components. HNR quantifies the relative amount of additive noise.

\subsection{Spectral Parameters}
     
\textbf{Alpha Ratio}: Ratio of the summed energy from 50-1000 Hz and 1-5 kHz.

\textbf{Hammarberg Index}: Ratio of the strongest energy peak in the 0-2 kHz region to the strongest peak in the 2–5 kHz region.

\textbf{Spectral Slope 0-500 Hz and 500-1500 Hz}: Linear regression slope of the logarithmic power spectrum within the two given bands.

\textbf{Formant 1, 2, and 3 relative energy}: The ratio of the energy of the spectral harmonic peak at the first, second, and third formant’s centre frequency to the energy of the spectral peak at F0.

\textbf{MFCC 1-4}: Mel-Frequency Cepstral Coefficients 1-4.

\textbf{Spectral flux}: Difference of the spectra of two consecutive frames.

\section{\textit{ImbOLL} and OLL parameters}
\label{sec:appendix-imboll}

Here, we show the experiments conducted to determine the optimal values for $\alpha$ and $\beta$ for our \textit{ImbOLL} function presented in equation \ref{eq:imboll}. We experiment with $\alpha \in \{1,1.5,2\}$ and $\beta \in \{0.5,1\}$. The results are shown in Table \ref{tab:imboll-exps}.

\begin{table}[t]
\centering
\begin{tabular}{@{}llr@{}}
\multicolumn{1}{c}{$\alpha$} & \multicolumn{1}{c}{$\beta$} & \multicolumn{1}{c}{Validation CCC($\uparrow$)} \\ \midrule
\multirow{2}{*}{1}   
                     & 0.5 & $\textbf{0.654} \pm \textbf{0.014}$ \\
                     & 1   & $0.653 \pm 0.022$ \\ \midrule
\multirow{2}{*}{1.5} 
                     & 0.5 &  $0.639 \pm 0.026$\\
                     & 1   & $0.516 \pm 0.024$ \\ \midrule
\multirow{2}{*}{2}   
                     & 0.5 & $0.610 \pm 0.038$ \\
                     & 1   & $0.422 \pm 0.024$ \\ \bottomrule
\end{tabular}
\caption{\textit{ImbOLL} experiments}
\label{tab:imboll-exps}
\end{table}

\begin{table}[t]
\centering
\begin{tabular}{ll}
\hline
$\alpha$ & Validation CCC($\uparrow$) \\ \hline
1                    & $\textbf{0.659} \pm \textbf{0.024}$        \\
1.5                  & $\textbf{0.655} \pm \textbf{0.012}$         \\
2                    & $0.645 \pm 0.014$         \\ \hline
\end{tabular}
\caption{OLL experiments}
\label{tab:oll-exps}
\end{table}

From Table \ref{tab:imboll-exps}, we see that $\alpha = 1$ gives the best results. Both $\beta = 0.5$ and $\beta=1$ give similar results on mean performance. However, $\beta = 0.5$ gives a lower standard deviation. Thus, we choose the value of $\alpha = 1$ and $\beta=0.5$ for training \textit{QuestMF}.

We also experiment with the hyperparameters of the OLL function presented in equation \ref{eq:oll}. We experiment with $\alpha \in \{1,1.5,2\}$, and the results are presented in Table \ref{tab:oll-exps}. From the table, we see that $\alpha = 1$ and $\alpha=1.5$ give good results. While $\alpha=1$ gives slightly better mean performance, $\alpha=1.5$ gives a lower standard deviation. So, we train \textit{QuestMF} with both $\alpha = 1$ and $\alpha = 1.5$. The results are presented in Appendix \ref{sec:appendix-ablation}. From that, we observe that OLL with $\alpha = 1$ gives the best result on the three-modality fused model. Thus, only this result is presented in Section \ref{sec:results}.

\section{Hyperparameter Details and Training Setup}
\label{sec:appendix-Exp-set-up}

This section presents the hyperparameters used in the single modality and fused two-modality and three-modality models. We use a maximum of $120$ participant dialogue turns for all our experiments. 

\textbf{Single Modality Encoder Models:} For the single modality models, we first experiment to find the best Bi-LSTM output dimensions. We experiment with hidden dimensions of $d \in {30,50,100}$. Due to the computational expenses, we only experimented with the text encoder model and extended the same output dimensions to audio and video encoder models. The results of this experiment are presented in Table \ref{tab:hyperparam_questmf_imboll} for the \textit{QuestMF} framework trained with the \textit{ImbOLL} function. The results of \textit{QuestMF} framework trained with OLL with $\alpha = 1$ are presented in Table \ref{tab:hyperparam_questmf_oll1} and with OLL with $\alpha = 1.5$ are shown in Table \ref{tab:hyperparam_questmf_oll15}. The results of \textit{QuestMF} framework trained with the MSE loss function are presented in Table \ref{tab:hyperparam_questmf_mse} and results of the \textit{Total} framework are presented in Table \ref{tab:hyperparam_total}. The tables show that a Bi-LSTM hidden dimension of $50$ works best for all frameworks. For the multihead attention layer, we use $4$ attention heads and a dropout of $0.5$ for the text encoder model for all frameworks. For the audio and video encoder models, we use two multihead attention layers with $4$ attention heads and a dropout of $0.2$ in all frameworks. The MLP in all single modality encoder models and frameworks consists of two linear layers with a hidden dimension of $256$, and the ReLU activation function connects the linear layers. A dropout of $0.2$ is applied before each linear layer.

For the training of the single modality encoders, we use a learning rate of $5 \times 10^{-4}$ with AdamW optimiser and a batch size of $10$ for all modalities. During training, we freeze the turn encoders and only train the Bi-LSTM layer, attention layer, and MLP. Since the text models fit faster, we train them for $20$ epochs. Meanwhile, we train the audio and video models for $50$ epochs. We select the model checkpoint with the lowest validation loss for further modality fusion training. To evaluate the models on the depression severity score prediction task, we select the checkpoint with the best validation CCC.

\begin{table}[t]
\centering
\begin{tabular}{ll}
\hline
Output Dimension & Validation CCC($\uparrow$) \\ \hline
30                    & $0.639 \pm 0.027$        \\
50                    & $\textbf{0.654} \pm \textbf{0.014}$         \\
100                   & $0.647 \pm 0.004$         \\ \hline
\end{tabular}
\caption{Results of experiments with LSTM output dimension for \textit{QuestMF} Framework trained with \textit{ImbOLL} function}
\label{tab:hyperparam_questmf_imboll}
\end{table}

\begin{table}[t]
\centering
\begin{tabular}{ll}
\hline
Output Dimension & Validation CCC($\uparrow$) \\ \hline
30                    & $0.622 \pm 0.047$        \\
50                    & $\textbf{0.659} \pm \textbf{0.024}$         \\
100                   & $0.639 \pm 0.038$         \\ \hline
\end{tabular}
\caption{Results of experiments with LSTM output dimension for \textit{QuestMF} Framework trained with OLL function with $\alpha = 1$}
\label{tab:hyperparam_questmf_oll1}
\end{table}

\begin{table}[t]
\centering
\begin{tabular}{ll}
\hline
Output Dimension & Validation CCC($\uparrow$) \\ \hline
30                    & $0.609 \pm 0.028$        \\
50                    & $\textbf{0.655} \pm \textbf{0.012}$         \\
100                   & $0.616 \pm 0.015$         \\ \hline
\end{tabular}
\caption{Results of experiments with LSTM output dimension for \textit{QuestMF} Framework trained with OLL function with $\alpha = 1.5$}
\label{tab:hyperparam_questmf_oll15}
\end{table}

\begin{table}[t]
\centering
\begin{tabular}{ll}
\hline
Output Dimension & Validation CCC($\uparrow$) \\ \hline
30                    & $0.554 \pm 0.022$         \\
50                    & $\textbf{0.632} \pm \textbf{0.024}$         \\
100                   & $0.602 \pm 0.008$         \\ \hline
\end{tabular}
\caption{Results of experiments with LSTM output dimension for \textit{QuestMF} Framework trained with MSE loss function}
\label{tab:hyperparam_questmf_mse}
\end{table}

\begin{table}[!th]
\centering
\begin{tabular}{ll}
\hline
Output Dimension & Validation CCC($\uparrow$) \\ \hline
30                    & $0.588 \pm 0.015$        \\
50                    & $\textbf{0.614} \pm \textbf{0.010}$         \\
100                   & $0.610 \pm 0.012$         \\ \hline
\end{tabular}
\caption{Results of experiments with LSTM output dimension for \textit{Total} Framework}
\label{tab:hyperparam_total}
\end{table}

\textbf{Two-Modality Fused Models:} We follow the architecture shown in Figure \ref{fig:two-mod-fusion} for the fusion of two modalities. We use $4$ heads and a dropout of $0.8$ for multihead cross-attention and self-attention layers. We use the very high dropout to reduce overfitting due to the small size of the training dataset. The MLP consists of two linear layers with a hidden dimension of $256$. The linear layers are connected through the ReLU activation function. We apply a dropout of $0.8$ before the first linear layer and a dropout of $0.5$ before the last linear layer. We use a smaller dropout before the last linear layer to avoid underfitting.

For the training of a two-modality fusion encoder, we use a learning rate of $5 \times 10^{-4}$ with AdamW optimiser and a batch size of $10$. If the text modality is involved in the two-modality fusion model, we freeze the weights from the text encoder model while we train the weights in the audio or video encoder models during the fusion. This is because the text model fits the data quickly, so training the weights of the other model with the frozen text model helps information alignment across the modalities and improves their encodings. Another reason is that training the parameters of all single modality encoder models with the small training set would increase the chances of overfitting. In addition to this, we train the cross-attention layers and self-attention layers used for fusion and the MLP. We train the models for $20$ epochs. We select the model checkpoint with the best validation CCC.

\textbf{Three-Modality Fused Models:} We follow the architecture shown in Figure \ref{fig:three-mod-fusion} for the fusion of three modalities. We use $4$ heads and a dropout of $0.8$ in multihead cross-attention and self-attention layers. We use an MLP of two linear layers with a hidden dimension of $256$. The linear layers are connected through the ReLU activation function. We apply a dropout of $0.8$ before the first linear layer and a dropout of $0.5$ before the last linear layer.

For the training of the three-modality fused model, we use a learning rate of $5 \times 10^{-4}$ with AdamW optimiser and a batch size of $10$. While training three-modality fused models, we freeze the weights from the text model and train the weights of the audio and video models along with the cross-attention and self-attention layers used for fusion, and the MLP. We train the fusion for $20$ epochs. We select the model checkpoint with the best validation CCC.

\section{Ablation Details}
\label{sec:appendix-ablation}
Here, we show more detailed results of our ablation study to observe the performance of single modality models, two-modality fused models, and the three-modality fused model with the frameworks. Here, we show the RMSE and MAE along with CCC results. We also show the standard deviation along with the mean for the three different seed runs. For \textit{QuestMF} (OLL), we have two different frameworks here:

\textbf{\textit{QuestMF (OLL-1)}}: We train the \textit{QuestMF} framework with OLL function with $\alpha = 1$.

\textbf{\textit{QuestMF (OLL-1.5)}}: We train the \textit{QuestMF} framework with OLL function with $\alpha = 1.5$.

We present the results in Table \ref{tab:ablation-results-table-app}.

\begin{table*}[t]
\centering
\begin{tabular}{@{}llrrr@{}}
\toprule
\multicolumn{1}{c}{Modalities} & \multicolumn{1}{c}{Framework} & \multicolumn{1}{c}{CCC($\uparrow$)} & \multicolumn{1}{c}{RMSE($\downarrow$)} & \multicolumn{1}{c}{MAE($\downarrow$)} \\ \midrule
\multirow{4}{*}{Text}                 & \textit{Total}            & $0.591 \pm 0.031$ & $5.51 \pm 0.38$ & $4.37 \pm 0.30$ \\
                                      & \textit{QuestMF} (MSE)    & $0.593 \pm 0.020$ & $5.52 \pm 0.09$ & $4.33 \pm 0.06$ \\
                                      & \textit{QuestMF} (OLL-1)    & $0.592 \pm 0.025$ & $5.77 \pm 0.31$ & $4.53 \pm 0.22$ \\
                                      & \textit{QuestMF} (OLL-1.5)    & $\textbf{0.616} \pm \textbf{0.019}$ & $\textbf{5.22} \pm \textbf{0.07}$ & $\textbf{4.02} \pm \textbf{0.07}$ \\
                                      & \textit{QuestMF} (\textit{ImbOLL}) & $0.615 \pm 0.031$ & $5.71 \pm 0.25$ & $4.36 \pm 0.26$  \\ \midrule
\multirow{4}{*}{Audio}                & \textit{Total}            & $0.212 \pm 0.017$ & $6.41 \pm 0.15$ & $5.27 \pm 0.09$ \\
                                      & \textit{QuestMF} (MSE)    & $0.239 \pm 0.012$ & $\textbf{6.35} \pm \textbf{0.09}$ & $\textbf{5.14} \pm \textbf{0.03}$ \\
                                      & \textit{QuestMF} (OLL-1)    & $0.264 \pm 0.008$ & $6.96 \pm 0.46$ & $5.42 \pm 0.20$ \\
                                      & \textit{QuestMF} (OLL-1.5)    & $0.256 \pm 0.023$ & $6.90 \pm 0.21$ & $5.36 \pm 0.14$ \\
                                      & \textit{QuestMF} (\textit{ImbOLL}) & $\textbf{0.273} \pm \textbf{0.021}$ & $6.67 \pm 0.11$ & $5.32 \pm 0.05$ \\ \midrule
\multirow{4}{*}{Video}                & \textit{Total}            & $-0.067 \pm 0.009$ & $8.21 \pm 0.07$ & $6.63 \pm 0.04$ \\
                                      & \textit{QuestMF} (MSE)    & $-0.075 \pm 0.007$ & $7.97 \pm 0.05$ & $6.46 \pm 0.04$ \\
                                      & \textit{QuestMF} (OLL-1)    & $\textbf{-0.041} \pm \textbf{0.015}$ & $7.91 \pm 0.20$ & $6.44 \pm 0.18$ \\
                                      & \textit{QuestMF} (OLL-1.5)    & $-0.044 \pm 0.026$ & $\textbf{7.79} \pm \textbf{0.19}$ & $\textbf{6.33} \pm \textbf{0.16}$ \\
                                      & \textit{QuestMF} (\textit{ImbOLL}) & $-0.052 \pm 0.028$ & $7.89 \pm 0.12$ & $6.44 \pm 0.11$ \\ \midrule
\multirow{4}{*}{Text + Audio}         & \textit{Total}            & $0.607 \pm 0.020$ & $\textbf{5.27} \pm \textbf{0.28}$ & $\textbf{4.12} \pm \textbf{0.23}$ \\
                                      & \textit{QuestMF} (MSE)    & $0.618 \pm 0.017$ & $5.61 \pm 0.31$ & $4.42 \pm 0.18$ \\
                                      & \textit{QuestMF} (OLL-1)    & $0.628 \pm 0.013$ & $5.44 \pm 0.04$ & $4.17 \pm 0.04$ \\
                                      & \textit{QuestMF} (OLL-1.5)    & $0.622 \pm 0.004$ & $5.53 \pm 0.07$ & $4.27 \pm 0.06$ \\
                                      & \textit{QuestMF} (\textit{ImbOLL}) & $\textbf{0.643} \pm \textbf{0.024}$ & $5.48 \pm 0.27$ & $4.21 \pm 0.14$ \\ \midrule
\multirow{4}{*}{Text + Video}         & \textit{Total}            & $0.610 \pm 0.008$ & $\textbf{5.13} \pm \textbf{0.19}$ & $4.02 \pm 0.17$ \\
                                      & \textit{QuestMF} (MSE)    & $0.627 \pm 0.021$ & $5.19 \pm 0.16$ & $3.99 \pm 0.05$ \\
                                      & \textit{QuestMF} (OLL-1)    & $0.628 \pm 0.006$ & $5.34 \pm 0.07$ & $4.04 \pm 0.20$ \\
                                      & \textit{QuestMF} (OLL-1.5)    & $0.630 \pm 0.039$ & $5.23 \pm 0.31$ & $4.05 \pm 0.30$ \\
                                      & \textit{QuestMF} (\textit{ImbOLL}) & $\textbf{0.659} \pm \textbf{0.018}$ & $5.22 \pm 0.09$ & $\textbf{3.92} \pm \textbf{0.08}$ \\ \midrule
\multirow{4}{*}{Audio + Video}        & \textit{Total}            & $0.058 \pm 0.030$ & $7.35 \pm 0.10$ & $5.86 \pm 0.11$ \\
                                      & \textit{QuestMF} (MSE)    & $0.070 \pm 0.011$ & $7.35 \pm 0.21$ & $5.81 \pm 0.16$ \\
                                      & \textit{QuestMF} (OLL-1)    & $0.139 \pm 0.016$ & $7.07 \pm 0.11$ & $5.69 \pm 0.08$ \\
                                      & \textit{QuestMF} (OLL-1.5)    & $0.080 \pm 0.047$ & $7.42 \pm 0.19$ & $6.02 \pm 0.20$ \\
                                      & \textit{QuestMF} (\textit{ImbOLL}) & $\textbf{0.159} \pm \textbf{0.039}$ & $\textbf{7.03} \pm \textbf{0.18}$ & $\textbf{5.67} \pm \textbf{0.25}$ \\ \midrule
\multirow{4}{*}{Text + Audio + Video} & \textit{Total}            & $0.602 \pm 0.015$ & $\textbf{5.10} \pm \textbf{0.08}$ & $3.99 \pm 0.05$ \\
                                      & \textit{QuestMF} (MSE)    & $0.602 \pm 0.025$ & $5.36 \pm 0.24$ & $4.21 \pm 0.12$ \\
                                      & \textit{QuestMF} (OLL-1)    & $0.640 \pm 0.018$ & $5.14 \pm 0.05$ & $\textbf{3.88} \pm \textbf{0.01}$ \\
                                      & \textit{QuestMF} (OLL-1.5)    & $0.599 \pm 0.023$ & $5.30 \pm 0.22$ & $4.06 \pm 0.25$ \\
                                      & \textit{QuestMF} (\textit{ImbOLL}) & $\textbf{0.662} \pm \textbf{0.022}$ & $5.25 \pm 0.08$ & $3.95 \pm 0.13$ \\ \bottomrule
\end{tabular}
\caption{Ablation results for using different combinations of modalities with different frameworks. The CCC, RMSE and MAE scores presented are the mean and standard deviation over $3$ different seed runs.}
\label{tab:ablation-results-table-app}
\end{table*}

\end{document}